\documentclass[fleqn,10pt]{SelfArx} 

\usepackage[english]{babel} 

\usepackage{listings}
\usepackage{xcolor}

\definecolor{codegreen}{rgb}{0,0.6,0}
\definecolor{codegray}{rgb}{0.5,0.5,0.5}
\definecolor{codepurple}{rgb}{0.58,0,0.82}
\definecolor{backcolour}{rgb}{0.95,0.95,0.92}

\lstdefinestyle{mystyle}{
    backgroundcolor=\color{backcolour},   
    commentstyle=\color{codegreen},
    keywordstyle=\color{magenta},
    numberstyle=\tiny\color{codegray},
    stringstyle=\color{codepurple},
    basicstyle=\ttfamily\footnotesize,
    breakatwhitespace=false,         
    breaklines=true,                 
    captionpos=b,                    
    keepspaces=true,                 
    numbers=left,                    
    numbersep=5pt,                  
    showspaces=false,                
    showstringspaces=false,
    showtabs=false,
    tabsize=2
}
\usepackage{pgfplots}
\pgfplotsset{width=8cm, compat=1.17}

\usepackage[style=numeric,sorting=none,backend=biber]{biblatex}
\usepackage{csquotes}

\usepackage{enumitem}

\usepackage{pgfplots}
\pgfplotsset{width=8cm}

\definecolor{color1}{RGB}{0,0,90} 
\definecolor{color2}{RGB}{0,20,20} 

\usepackage{hyperref} 

\hypersetup{
        hidelinks,
        colorlinks,
        breaklinks=true,
        urlcolor=color2,
        citecolor=color1,
        linkcolor=color1,
        bookmarksopen=false,
        pdftitle={Enhancing Formal Theorem Proving: A Comprehensive Dataset for Training AI Models on Coq Code},
        pdfauthor={Andreas Florath},
}

\JournalInfo{V2} 
\Archive{\ } 

\PaperTitle{Enhancing Formal Theorem Proving: A Comprehensive Dataset for Training AI Models on Coq Code}

\Authors{Andreas Florath\textsuperscript{1}*} 
\affiliation{\textsuperscript{1}\textit{flonatel GmbH \& Co.~KG, Aachen, Germany}}
\affiliation{*\textbf{Corresponding author}: andreas@florath.net}

\newcommand{\keywordname}{Keywords}

\Abstract{%
In the realm of formal theorem proving, the Coq proof assistant stands
out for its rigorous approach to verifying mathematical assertions and
software correctness.  Despite the advances in artificial intelligence
and machine learning, the specialized nature of Coq syntax and
semantics poses unique challenges for Large Language Models (LLMs).
Addressing this gap, we present a comprehensive dataset specifically
designed to enhance LLMs' proficiency in interpreting and generating
Coq code.  This dataset, derived from a collection of over
10,000 Coq source files, encompasses a wide array of propositions,
proofs, and definitions, enriched with metadata including source
references and licensing information.  Our primary aim is to
facilitate the development of LLMs capable of generating syntactically
correct and semantically meaningful Coq constructs, thereby advancing
the frontier of automated theorem proving.

Initial experiments with this dataset have showcased its significant
potential; models trained on this data exhibited enhanced accuracy in
Coq code generation.  Notably, a particular experiment revealed that a
fine-tuned LLM was capable of generating 141 valid proofs for a basic
lemma, highlighting the dataset's utility in facilitating the
discovery of diverse and valid proof strategies.  This paper discusses
the dataset's composition, the methodology behind its creation, and
the implications of our findings for the future of machine learning in
formal verification.

The dataset is accessible for further research and exploration:\newline
\url{https://huggingface.co/datasets/florath/coq-facts-props-proofs-gen0-v1}
}

\begin{document}

\maketitle 

\tableofcontents 

\thispagestyle{empty} 


\section{Introduction}

In the exploration of Large Language Models (LLMs) for code
optimization~\cite{florath2024llm}, two significant limitations were
identified:

\begin{itemize}[noitemsep,topsep=0pt,parsep=0pt,partopsep=0pt]
\item The dependency on human interaction impedes the model's ability
  to function autonomously, limiting its applicability to extensive
  source code collections and automation processes.
\item The indefinite nature of optimization completion, where a
  considerable portion of time is allocated to verification rather
  than the optimization process itself. The measurement of
  optimization efficacy remains a challenge.
\end{itemize}

The adoption of formal mathematical proofs presents a logical
advancement for overcoming the second limitation.  Through formal
proof assistants like Coq~\cite{CoqHome}, Lean~\cite{WebLean}, or
Isabelle~\cite{WebIsabelle}, the verification of propositions (such as
lemmas or theorems) becomes definitive.  Once verified, the
proposition is conclusively validated, eliminating the need for
further evaluation.

This approach advocates for focusing on domains akin to programming,
yet characterized by determinate termination states.  The development
of a system, potentially using agent-based models, is proposed.  Such
a system could subsequently be applied to the autonomous optimization
of source code, thereby resolving the identified challenges.

\section{Objectives}

This research endeavors to advance the integration of machine learning
and artificial intelligence within the realm of formal theorem
proving, emphasizing the Coq Proof Assistant. By developing a
dedicated dataset, this work aims to refine ML models, notably
enhancing LLMs' capabilities in processing and generating Coq
code. The objectives are meticulously outlined to encompass:

\begin{description}[style=unboxed, leftmargin=0cm, itemsep=0ex]
\item[Enhance Syntax and Semantic Comprehension:] Enhancing LLMs'
  proficiency in interpreting and generating Coq code by providing a
  comprehensive dataset, thereby facilitating a deeper comprehension
  of Coq's syntax, mathematical logic, and proof strategies.
\item[Enable Autonomous Content Generation:] Empowering LLMs to
  autonomously formulate mathematical definitions, lemmas, examples,
  and exercises, adjusting the complexity to bolster formal
  mathematics contributions.
\item[Optimize Coq Files for Machine Interaction:] Refining Coq
  codebases for improved machine interaction through simplification
  and standardization, aiming for broader application and usage.
\item[Facilitate Proof Generation:] Equipping LLMs with the necessary
  tools for autonomous proof generation, laying a foundation for
  innovative advancements in formal proofs.
\end{description}

The pursuit of these objectives is anticipated to elevate LLMs'
efficiency with Coq code, marking significant progress in automated
theorem proving and broadening the horizons for formal mathematics and
computer science research.

\section{Prior Art}\label{sec:PriorArt}

A singular comprehensive dataset, \textsl{The Stack v2}, has been
identified amidst extensive research efforts as encompassing a diverse
and extensive collection of Coq source
code~\cite{lozhkov2024starcoder}.  Hosting over 150,000 files, with
nearly 80,000 under a permissive license, the dataset stands out by
providing identifiers for source code retrieval from S3 storage rather
than including the code directly.  Unprocessed raw data
constitutes the dataset's format, presenting each file in a single
row.  Notably, precise and detailed license documentation is provided
for each file, an approach mirrored in the dataset discussed herein.

On Huggingface~\cite{HuggingfaceDatasets}, four additional datasets
containing Coq source code were found.  Two of these datasets
comprised entire Coq files within single rows, leading to impractical
usability due to excessively large row sizes, with the largest
containing over 6 million
characters~\cite{HuggingfaceDatasetCoqGithubScrape,
  HuggingfaceDatasetCoqTrain}.  Although these collections were
sizable, the licensing terms were inadequately addressed, mixing data
from various repositories under different licenses without proper
license adherence.  Queries regarding licensing prompted the removal
of both datasets.


CoqGym~\cite{yang2019coqgym} presents another notable attempt,
offering a substantial collection under the Creative Commons
Attribution 2.0 Generic License~\cite{CC2Attr2Gen}, which is
incompatible with the licenses of the included Coq source
code~\cite{CCBYSA4GPL3Compatibility}.  The issue of license
compatibility remains unresolved~\cite{CoqGymLicenseIssue}.
Furthermore, CoqGym duplicated content from other projects into its
repository, resulting in a dataset that is now outdated by five years.

The dataset "coq\_code"~\cite{HuggingfaceDatasetJBBCoqCode} on
Huggingface, though adhering to a step-by-step format (including
hypothesis, goal, and tactic), is limited, containing fewer than
25,000 entries.  Its formatting is suboptimal, with data merged into a
single text column and separated by special tags.

In parallel efforts to utilize machine learning for enhancing formal
proving in Coq, research has been conducted on the automation of lemma
name generation, leveraging a dataset constructed from approximately
450 Coq source files from the math-comp project.  This dataset, aimed
at producing AST and token files through preprocessing, encountered
challenges in data bloat and clarity, raising questions on its
efficacy for LLM training or fine-tuning.  To date, there's no
documented success in employing this specific dataset for LLM
enhancement~\cite{NieETAL20Roosterize, RepoMathCompCorpus}.  Another
effort was formatting Coq code using language
models.~\cite{nie2020learning}

No datasets containing Coq source code were found on Kaggle at the
time of this writing.~\cite{KaggleSearchForCoq}

Against the backdrop of these endeavors, the dataset presented in this
paper distinguishes itself through a unique combination of scale,
organization, and focus on formal theorem proving.  Unlike previously
mentioned datasets, which either offer raw, unprocessed files or are
constrained by licensing and formatting issues, this dataset provides
a curated and processed collection of Coq code.

Two recent publications, although not directly related to the dataset
focus of this paper, share similar approaches or motivations:

An approach is described where a large-scale, graph-based dataset and
a graph neural network are employed to dynamically integrate and
leverage the hierarchical structure of definitions, theorems, and
proofs within Coq. This method significantly enhances AI agents'
capability to adapt to new mathematical concepts and lemmas not
encountered during training, presenting a critical advancement in the
automation of theorem proving~\cite{rute2024graph2tac}.

A novel methodology employing Monte Carlo Tree Search (MCTS) to guide
LLMs for the generation of verified programs in Dafny, Lean, and Coq,
named VMCTS, enhances synthesis capabilities by incorporating verifier
feedback directly into the search algorithm, showcasing its efficiency
by solving complex verification problems in notably shorter times
compared to base models and even rivaling ChatGPT4's augmented
capabilities~\cite{brandfonbrener2024verified}.

\section{Data Sources}

The Coq source files for the datasets were meticulously collected from
a diverse array of sources across the internet, focusing on
repositories that are pivotal within the Coq community and cover a
broad spectrum of mathematical and computational theories.  These
sources encompass a range of categories, including foundational
libraries, formalized mathematical theorems, computer science
concepts, and algorithm implementations.

Foundational Libraries and Frameworks form the bedrock, with
repositories like the official Coq repository~\cite{RepoCoq},
math-comp (Mathematical Components)~\cite{RepoMathComp}, and Coq's
standard library extensions~\cite{RepoCoqExtLib}.  These are essential
for anyone working with Coq, offering basic definitions, theorems, and
tactics widely used in further Coq developments.

Formalized Mathematics and Theorem Proofs are represented through
collections such as GeoCoq (geometry)~\cite{RepoGeoCoq}, the formal
proofs of the Four Color Theorem~\cite{RepoTheFourColorTheorem}, and
various projects under the Coq-community umbrella focusing on specific
mathematical domains like algebra~\cite{RepoAlgebraTactics}, number
theory~\cite{RepoCoqPrime}, and logic~\cite{Repo100Theorems}. These
projects not only provide proofs of known theorems but also extend the
library of formalized mathematics accessible for Coq users.

Computer Science Theories and Algorithms feature prominently, with
projects like Verdi (for distributed systems
verification)~\cite{RepoVerdi}, the Iris project for concurrent
systems~\cite{RepoIris}, and various algorithm collections including
sorting, graph theory, and data structures.  These repositories are
crucial for researchers and practitioners interested in the formal
verification of software and algorithms.

The repositories were chosen for their quality, relevance to the Coq
community, and contribution to the ecosystem.  The collected datasets
aim to provide comprehensive coverage of the syntax and semantics
employed in Coq development, supporting the project's goal of
enhancing LLMs' understanding and generation capabilities with respect
to Coq code.  The datasets ensure a wide representation of the Coq
language's potential applications, from pure mathematics to computer
science.

\section{Licenses}

Addressing the complexities of licensing within the context of
aggregating datasets from various sources is a non-trivial
challenge.~\cite{lozhkov2024starcoder} The datasets compiled for
enhancing Large Language Models' (LLMs) comprehension and generation
of Coq code embody this challenge, as they amalgamate content from a
multitude of repositories, each governed by its unique license.  Given
the diverse origins of the Coq source files, the datasets do not
subscribe to a singular license.  Instead, each row in the facts and
proposition / proofs table link to the license table where for each
row the needed information can be found.

To comply with the stipulations of these licenses, especially those
like MIT which mandate the inclusion of original licensing and
authorship information, the dataset incorporates copies of the
original license files and, where available, the author files.  This
practice ensures adherence to the legal requirements of software
redistribution, particularly for open-source licenses that permit such
activities.

The compilation strictly omits libraries or files that lack an
explicit open-source license or are under a commercial license,
thereby ensuring that the dataset comprises only data that is legally
redistributable. This careful selection process is pivotal for
maintaining the integrity and legality of the datasets, facilitating
their use in research and development without infringing upon
copyright laws or license conditions.

The dataset encompasses a wide range of licenses, reflecting the
diversity of the Coq community and the broader open-source
ecosystem. Among these are:

\begin{itemize}[noitemsep,topsep=0pt,parsep=0pt,partopsep=0pt]
\item Apache License 2.0 (Apache-2.0)
\item BSD 2-Clause "Simplified" License (BSD-2-Clause)
\item BSD 3-Clause "New" or "Revised" License (BSD-3-Clause)
\item CEA CNRS Inria Logiciel Libre License, versions 1.0, 2.1
  (CECILL-1.0, CECILL-2.1), including its variants CECILL-B and
  CECILL-C for library and plugin distributions, respectively
\item GNU General Public License versions 2.0 only (GPL-2.0-only), 3.0
  only (GPL-3.0-only), and 3.0 or later (GPL-3.0-or-later)
\item GNU Lesser General Public License versions 2.1 only
  (LGPL-2.1-only), 2.1 or later (LGPL-2.1-or-later), 3.0 only
  (LGPL-3.0-only), and 3.0 or later (LGPL-3.0-or-later)
\item MIT License (MIT)
\item Mozilla Public License 2.0 (MPL-2.0)
\item UniMath License (specific to the UniMath library)
\end{itemize}

This approach ensures that the datasets not only respect the legal and
ethical considerations of software redistribution but also provide a
rich, legally compliant resource for advancing the capabilities of
LLMs in processing and generating Coq code.

\section{Dataset ``coq-facts-props-proofs''}

This dataset is comprised of three distinct tables:

\begin{enumerate}[itemsep=0ex]
\item Definitions or notations categorized as facts.
\item Theorems and lemmas, alongside their proofs, classified as
  propositions.
\item Licensing and repository information for each entry within the
  facts and propositions tables.
\end{enumerate}

License identification was conducted manually: a license hint within
the \texttt{Readme} file was prioritized, followed by the contents of
any \texttt{LICENSE} file.  Only repositories under open-source
licenses permitting redistribution were included.

The dataset exclusively features Coq source code files (\texttt{.v}
files), which were pre-processed using a customized OCaml parser to
separate Coq sentences, remove comments, and eliminate directives like
\texttt{\#global}.  This process also involved condensing multiple
consecutive whitespaces into a single space and deduplicating based on
facts and proposition/proof content rather than file origin.  The
preprocessing was purely done on parsing level, no evluation of the
Coq source code was done.  Consequently, some parts of the Coq code
may not evaluate or may not be compatible with the latest version of
Coq.

The \texttt{facts} table is one cornerstone of the dataset,
encompassing definitions or notations.  Each row within this table
represents a unique fact, identified by a Coq definition or notation.
These facts are detailed through several key columns:
\begin{description}[itemsep=0ex]
\item[fact] the fact itself, presented in Coq syntax
\item[imports] a list of imports, specifying the Coq modules and
  libraries required for the fact's context
\item[filename] the filename, indicating
  the source file from which the fact was extracted
\item[symbolic\_name] the symbolic name, providing a reference handle
  for the fact to the repository and license information.
\end{description}

The table \texttt{props-proofs} is the other key component of the
dataset.  The structure is very similar to the facts table, but
instead of using the \textbf{facts} column there are two columns
\textbf{proposition} and \textbf{proof}.

The "info" table within our dataset acts as a vital link between the
\textbf{symbolic\_name} and its corresponding repository, enriched with
precise licensing information. It is comprised of four columns:
\begin{description}[itemsep=0ex]
\item[symbolic\_name] serving as a unique identifier correlating to
  entries within the "facts" and "props-proofs" tables
\item[url] providing the repository's location which hosts the source
  Coq files
\item[hexsha] representing the Git SHA of the last commit at
  the time the repository was checked out, offering a snapshot for
  reproducibility and version tracking
\item [spdx-id] detailing the license under which the repository's
  content is distributed, in alignment with the Software Package Data
  Exchange (SPDX) identifiers.
\end{description}


The dataset is accessible on huggingface:~\cite{DatasetOnHuggingface}.

\section{Statistics}

\subsection{info.parquet}

The \texttt{info.parquet} table comprises 142 rows, each representing
a repository.  The distribution of licenses across these repositories
is outlined below:

\begin{center}
\begin{tabular}{l|r||l|r}
\textbf{License} & \textbf{Count} & \textbf{License} & \textbf{Count}\\
\hline  
MIT                  &43 & Apache-2.0           & 3\\
LGPL-2.1-only        &29 & MPL-2.0              & 3\\
LGPL-2.1-or-later    &12 & GPL-3.0-only         & 3\\
CECILL-B             & 9 & GPL-3.0-or-later     & 3\\
CECILL-1.0           & 7 & BSD-3-Clause         & 2\\
LGPL-3.0-only        & 7 & CECILL-2.1           & 2\\
LGPL-3.0-or-later    & 6 & GPL-2.0-only         & 1\\
CECILL-C             & 6 & CECILL-2.0           & 1\\
BSD-2-Clause         & 4 & UniMath              & 1
\end{tabular}
\end{center}

\subsection{facts.parquet}

Data pertaining to the facts.parquet table is provided below, with
measurements based on character count:

\begin{center}
\begin{tabular}{|l|r|}
\hline
Columns & 4\\
Rows & 103,446\\
Shortest fact&12\\
Longest fact & 37,630\\
Mean length & 132.26\\
Standard deviation & 359.47\\
\hline
\end{tabular}
\end{center}

\subsection{props-proof.parquet}

Details regarding the props-proof.parquet table are summarized below,
with lengths measured in characters:

\begin{center}
\begin{tabular}{|l|r|}
\hline
Columns & 5\\
Rows & 166,035\\
Shortest proposition & 13\\
Longest proposition & 7400\\
Mean length proposition & 104.05\\
Standard deviation proposition & 97.65\\
Shortest proof & 11\\
Longest proof & 177585\\
Mean length proof & 347.88\\
Standard deviation proof & 1290.80\\
\hline
\end{tabular}
\end{center}

\begin{figure}[ht]\centering
\includegraphics[width=\linewidth]{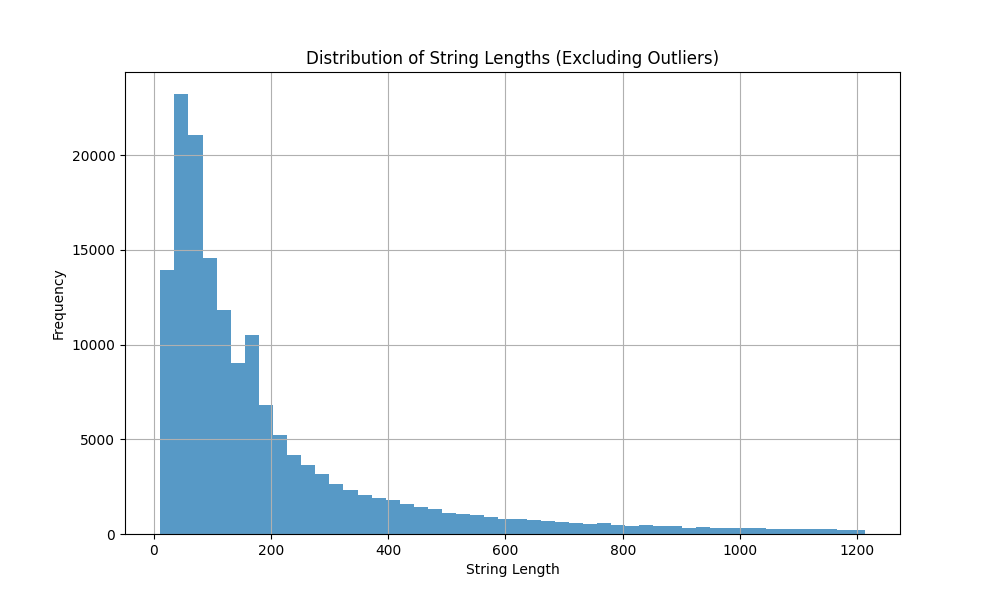}
\caption{Deviation of length of proofs for the 0.95 percentile}
\label{ProofDeviation}
\end{figure}

Observations indicate high standard deviations, attributed to the
presence of a few exceptionally long facts, propositions, and
proofs.  The deviation pattern when excluding the top 5\% of
length can be seen in figure \ref{ProofDeviation}.

\section{Experiments}

In this section, we explore one of many possible applications of the
dataset through the fine-tuning of an existing base model,
Mistral-7b~\cite{jiang2023mistral}.  This exercise is meant to serve
as an illustration of the dataset's potential rather than a
comprehensive or central focus of the paper.  Our intention is to
demonstrate, via selected examples, how the dataset can be utilized to
potentially enhance LLM's understanding and generation of Coq code.

The fine-tuning process, performed on an NVidia A30 GPU across
approximately seven days, involved adapting the model to better handle
Coq syntax and logic as represented in the dataset.  Every three hours
a snapshot of the model was generated.  It's important to note that
while the model's performance post fine-tuning provides insights into
the dataset's utility, it represents only one of many possible
evaluation metrics.

The model's output underwent evaluation at a temperature setting of
$0.4$ across different snapshots using \texttt{coqc} or
\texttt{coqide}.  We curated the output for readability, truncating
responses at logical endpoints such as \texttt{Qed.}, to focus on the
model's capability to produce syntactically and logically coherent Coq
constructs.  These choices were guided by the goal of assessing the
model's ability to generate syntactically and logically correct Coq
code, underlining the qualitative rather than quantitative nature of
this experiment.  A version of the model which was trained only using
Coq code with permissive licenses is publicity
available~\cite{ModelOnHuggingface}.

Additionally, we made prompt adjustments to encourage Coq-specific
responses from the different models, indicating the necessity of
tailored inputs for optimal output in domain-specific tasks.  The
comparison of the fine-tuned model against several prominent LLMs
provides a broader context for evaluating the dataset's impact on
enhancing Coq code generation capabilities, albeit this comparison is
illustrative of the dataset's potential rather than an exhaustive
evaluation of its efficacy.

The models under observation and for comparison:
\begin{description}[style=unboxed, leftmargin=0cm, itemsep=0ex]
\item[CoqLLM-FineTuned] This model was fine-tuned with the
  complete dataset described in this paper, specifically designed
  to enhance its proficiency in interpreting and generating Coq code
  and serves as the experiment to show the usefulness of the dataset.
\item[Mistral-7b-Instruct-0.2] Based on the Mistral-7b architecture,
  this model leverages instructional data to guide its responses and
  programming language understanding.~\cite{jiang2023mistral}
\item[Starcoder2-15b] Starcoder2-15b has been trained on
  over 600 different programming languages, including Coq, providing
  it with a broad syntax and semantic understanding across a wide
  array of languages.~\cite{lozhkov2024starcoder}
\item[Google Gemini] This publicly available chat model from Google
  demonstrates capabilities in natural language processing and
  understanding, applied across various contexts, including
  programming.~\cite{geminiteam2023gemini}
\item[OpenAI ChatGPT 4] As OpenAI's publicly available chat model,
  ChatGPT 4 showcases advancements in language models' ability to
  engage in detailed conversations and generate code
  snippets.~\cite{openai2024gpt4}
\end{description}

\subsection{Experiment 1: n = n + 0}\label{sec:Experiment1}

\subsubsection{Prompt and Reference Proof}

For this experiment, the lemma tested was as follows:
\begin{lstlisting}
Lemma plus_n_O : forall n:nat, n = n + 0.
\end{lstlisting}

The reference proof contained within the training data is
straightforward~\cite{RepoGeoCoq}:
\begin{lstlisting}
Proof.
  induction n; trivial.
Defined.
\end{lstlisting}

\subsubsection{Model Responses}

Among the 563 responses generated that began with \texttt{Proof.}, 141
were identified as valid (see
section \ref{sec:ManyWaysToProofTheLemma}), demonstrating
the model's adeptness not only in understanding Coq syntax but also in
navigating its semantic landscape to reach valid conclusions through
various methods.

Notably, the variety of proofs highlights the LLM's capacity to
utilize a broad spectrum of Coq's proof strategies, ranging from
direct application of arithmetic simplification (\texttt{auto with
  arith.}) to structural induction and recursive definitions
(\texttt{induction n as [| n IHn].}).  This diversity not only
showcases the potential of LLMs in theorem proving but also suggests a
nuanced understanding of the Coq proof assistant's capabilities,
opening new avenues for exploring automated theorem proving.

These findings are particularly significant as they suggest that LLMs,
when equipped with a well-curated dataset, can extend beyond mere
syntactic correctness to exhibit a deep comprehension of mathematical
logic and proof strategies.  This depth enables the generation of
multiple, distinctively valid approaches to proving a single
proposition, thereby enriching the repertoire of automated theorem
proving.

These implications reinforce the utility of specialized datasets in
enhancing the performance of LLMs within domain-specific tasks such as
theorem proving.

\subsubsection{Comparative Model Responses}

\begin{description}[style=unboxed, leftmargin=0cm, itemsep=0ex]
\item[Mistral-7b-Instruct] Responded in a non-Coq language and failed
  to generate a valid proof even after prompt adaptation.
\item[ChatGPT 4] Although replying in Coq, the proof offered was
  incorrect.
\item[Google Gemini] Required prompt modification before producing a
  correct proof. 
\item[Starcoder2-15b] Did not provide any proof, despite being
  prompted.
\end{description}

\subsubsection{Discussion}

This experiment highlights the CoqLLM-FineTuned model's superior
capability in producing correct Coq proofs that were not part of its
training set, distinguishing it from other models, including those of
similar size and significantly larger ones like ChatGPT 4 or Google
Gemini.  The model not only demonstrated its understanding of Coq
syntax and logic but also its ability to creatively solve problems
without directly reproducing training data.

\subsection{Experiment 2: 7 + 3 = 10}

\subsubsection{Prompt and Theoretical Proof}

The prompt for this experiment was:
\begin{lstlisting}
Lemma ex1: 7 + 3 = 10.
\end{lstlisting}

Notably, this specific lemma did not exist within the training
dataset.  However, a theoretically valid proof employing basic
reflexivity is suggested:

\begin{lstlisting}
Proof.
  reflexivity.
Qed.
\end{lstlisting}

\subsubsection{Model's Response}

Remarkably, the CoqLLM-FineTuned model independently arrived with most
responses at the same proof as the one proposed, successfully
utilizing the \texttt{reflexivity} tactic.

\subsubsection{Comparison with Other Models}

\begin{description}[style=unboxed, leftmargin=0cm, itemsep=0ex]
\item[Mistral-7b-Instruct] Failed to provide a valid Coq proof,
  responding inappropriately and deviating significantly from the
  prompt.
\item[ChatGPT 4, Google Gemini, and Starcoder2-15b] Each of these
  models managed to produce valid proofs, indicating a general
  competence in handling straightforward arithmetic propositions in
  Coq.
\end{description}

\subsubsection{Discussion}

This experiment underscores the performance of the CoqLLM-FineTuned
model in generating a valid proof for a proposition not present in its
training set, further exemplifying its advanced reasoning
capabilities.  Unlike the Mistral-7b-Instruct model, which failed to
generate a correct response, the CoqLLM-FineTuned, alongside other
prominent models like ChatGPT 4, Google Gemini, and Starcoder2-15b,
demonstrated proficiency in Coq syntax and logical reasoning.

\subsection{Experiment 3: S (m * n) = m * n + n.}

\subsubsection{Prompt and Challenge}

The lemma explored in this experiment was as follows:
\begin{lstlisting}
Lemma mult_S : forall m n : nat, S (m * n) = m * n + n.
\end{lstlisting}

Intentionally erroneous, this lemma serves to test the LLMs' ability
to recognize or question the validity of a proposition, essentially
assigning them an impossible task.

\subsubsection{Discussion}

Despite the intrinsic fallacy in the lemma, all tested models,
including Mistral-7b-Instruct, ChatGPT 4, Google Gemini, Starcoder2,
and CoqLLM-FineTuned, endeavored to construct a proof without
indicating any recognition of the proposition's incorrectness.  This
uniform approach across diverse models reveals a critical area for
future enhancement in LLMs' capabilities: the detection of inherently
flawed or unsolvable problems.

\section{Results}
The fine-tuning of a Large Language Model (LLM) with the Coq dataset
demonstrated promising outcomes, with the model generating outputs
with a high probability exclusively in Coq syntax.  This specificity
in output underscores the dataset's effectiveness in aligning the
trained model with the requirements of both agent systems and Coq
runtime environments, making it a preferred choice for these
applications.

The endeavor also revealed the feasibility of achieving significant
advancements in model performance with limited resources and within a
constrained timeframe.  The refined model showcased an ability to
produce insightful, Coq-compatible remarks, underscoring the potential
for further enhancing the efficiency of theorem proving in Coq.

Moreover, the careful curation, cleanup, and licensing of the dataset
not only facilitated this study but also ensure its utility for the
broader research community.  This resource is poised to contribute to
the ongoing development of agents, marking a crucial step in the
journey towards more sophisticated and autonomous theorem proving
systems.

Building upon these achievements, the notable success in Experiment 1
(see section \ref{sec:Experiment1}), where the fine-tuned LLMs
generated 141 valid proofs for the proposition $n = n + 0$ opens a new
vista for the application of LLMs in generating valuable Coq source
code.  This accomplishment illustrates the models' capacity not only
to adhere to syntactic correctness but also to engage in creative
problem-solving within the Coq framework.  The presence of valid,
varied proofs further underscores the potential utility of LLMs as
tools for enriching and expanding Coq datasets with new, verified
source code.

\section{Outlook}

The successful fine-tuning of the Large Language Model (LLM) using the
Coq dataset opens up several promising avenues for future research and
application enhancements:

\begin{description}[style=unboxed, leftmargin=0cm, itemsep=0ex]
\item[Agent-based Application:] The dataset can serve as a training
  data for models for developing agents capable of interacting with,
  and reasoning about, Coq code.  This could significantly streamline
  processes in formal methods and theorem proving by providing
  automated assistance.
\item[Refining Prompts with the Dataset:] Utilizing the dataset to
  fine-tune prompts can enhance the generation of higher quality and
  more relevant content. This improvement can bolster the model's
  capacity to tackle intricate problem-solving and reasoning within
  formal verification's scope.
\item[Hypothesis, Goal, Tactic Approach:] Implementing a structured
  approach that defines hypotheses, sets goals, and employs tactics
  could further sophisticate the model's interaction with formal
  proofs.  This strategy could facilitate the development of more
  advanced models capable of autonomously devising and verifying
  proofs, thus pushing the boundaries of what's possible in automated
  theorem proving.
\item[Curating, Enhancing, and Enlarging the Coq Source Code Base
  Using LLMs:] The demonstrated capacity of LLMs to generate valid and
  diverse Coq proofs opens the door to utilizing these models for the
  curation, enhancement, and expansion of the Coq source code base.  By
  integrating LLM-generated code into the dataset, we can continuously
  improve the quality and breadth of Coq resources available to the
  research community, fostering a more dynamic and robust ecosystem
  for theorem proving and formal verification.
\end{description}

These directions not only underscore the potential for the dataset to
catalyze significant advancements in the field but also highlight the
broader implications for enhancing AI's role in formal methods and
beyond.

\section{Acknowledgment}
The author extend his sincere gratitude to Andrei Agapi for his
invaluable ideas and for conducting the initial review, which
significantly contributed to the enhancement of this paper.

``Stylish Article'' template \cite{StylishArticleLaTeXTemplate2022}
was used.

\onecolumn

\section{Appendix}

\subsection{141 Ways to Proof the Lemma}\label{sec:ManyWaysToProofTheLemma}

The validation was performed using \texttt{coqc} version 8.19.1.  In
addition to the proposition, the validated coq source file contained some
\texttt{Require Imports} as well as some notations like
\texttt{plus\_0\_l} or \texttt{plus\_comm}.

\begin{lstlisting}
Proof. auto with arith. Qed.
Proof. destruct n; simpl; auto. Qed.
Proof. induction n; simpl; auto. Qed.
Proof. intro n; apply eq_add_S; auto with arith. Qed.
Proof. intro n; destruct n as [|p]. - reflexivity. - simpl; auto. Qed.
Proof. intro n; elim n ; auto with arith. Qed.
Proof. intro n; elim n ; simpl; auto. Qed.
Proof. intro n; elim n using nat_rect. - reflexivity. - intros m Hrec; simpl; auto. Qed.
Proof. intro n; elim n using nat_rect. - reflexivity. - intros m Hrec; simpl; rewrite <- Hrec; auto with arith. Qed.
Proof. intro n; elim n using nat_rect. - reflexivity. - simpl. auto with arith. Qed.
Proof. intro n; elim n using nat_rect; auto. Qed.
Proof. intro n; elim n. - reflexivity. - intros m IHm. simpl. rewrite <- IHm. reflexivity. Qed.
Proof. intro n; elim n. - reflexivity. - intros p IHp; simpl; rewrite <- IHp; reflexivity. Qed.
Proof. intro n; elim n; auto with arith. Qed.
Proof. intro n; elim n; auto. Qed.
Proof. intro n; elim n; simpl. - reflexivity. - intros m IHm; rewrite <- IHm; reflexivity. Qed.
Proof. intro n; elim n; simpl. reflexivity. intros m IH. rewrite <- IH. auto with arith. Qed.
Proof. intro n; elim n; simpl. reflexivity. intros m IHm. rewrite <- IHm. reflexivity. Qed.
Proof. intro n; elim n; simpl; auto with arith. Qed.
Proof. intro n; elim n; simpl; auto. Qed.
Proof. intro n; induction n as [ | n IHn]. - reflexivity. - simpl; rewrite <- IHn; auto. Qed.
Proof. intro n; induction n as [ | p IHp]. - reflexivity. - simpl; rewrite <- IHp; reflexivity. Qed.
Proof. intro n; induction n as [| m IHm]; auto with arith. Qed.
Proof. intro n; induction n as [| n Hrecn]. - reflexivity. - simpl; rewrite <- Hrecn; reflexivity. Qed.
Proof. intro n; induction n as [| n IH]. - reflexivity. - simpl. apply f_equal. assumption. Qed.
Proof. intro n; induction n as [| n IH]. - reflexivity. - simpl. rewrite <- IH. reflexivity. Qed.
Proof. intro n; induction n as [| n IH]. - reflexivity. - simpl; auto. Qed.
Proof. intro n; induction n as [| n IH]. - reflexivity. - simpl; rewrite <- IH; reflexivity. Qed.
Proof. intro n; induction n as [| n IH]; auto. Qed.
Proof. intro n; induction n as [| n IH]; simpl. - reflexivity. - rewrite <- IH; ring. Qed.
Proof. intro n; induction n as [| n IH]; simpl; auto. Qed.
Proof. intro n; induction n as [| n IHn]. - reflexivity. - simpl. rewrite <- IHn. reflexivity. Qed.
Proof. intro n; induction n as [| n IHn]. - reflexivity. - simpl; auto. Qed.
Proof. intro n; induction n as [| n IHn]. - reflexivity. - simpl; now rewrite <- IHn. Qed.
Proof. intro n; induction n as [| n IHn]. - reflexivity. - simpl; rewrite <- (IHn); auto. Qed.
Proof. intro n; induction n as [| n IHn]. - reflexivity. - simpl; rewrite <- (plus_n_O n); reflexivity. Qed.
Proof. intro n; induction n as [| n IHn]. - reflexivity. - simpl; rewrite <- IHn at 1; reflexivity. Qed.
Proof. intro n; induction n as [| n IHn]. - reflexivity. - simpl; rewrite <- IHn; auto. Qed.
Proof. intro n; induction n as [| n IHn]. - reflexivity. - simpl; rewrite <- IHn; reflexivity. Qed.
Proof. intro n; induction n as [| n IHn]. - reflexivity. - simpl; rewrite IHn at 1; reflexivity. Qed.
Proof. intro n; induction n as [| p Hp]. - reflexivity. - simpl; rewrite <- Hp; reflexivity. Qed.
Proof. intro n; induction n as [| p Hp]. - simpl; auto with arith. - simpl; rewrite <- Hp; reflexivity. Qed.
Proof. intro n; induction n as [| p IHp]. - reflexivity. - simpl. rewrite <- (IHp). reflexivity. Qed.
Proof. intro n; induction n as [| p IHp]. - reflexivity. - simpl. rewrite <- IHp. reflexivity. Qed.
Proof. intro n; induction n as [| p IHp]. - reflexivity. - simpl; auto with arith. Qed.
Proof. intro n; induction n as [| p IHp]. - reflexivity. - simpl; auto. Qed.
Proof. intro n; induction n as [| p IHp]. - reflexivity. - simpl; rewrite -> IHp at 1; reflexivity. Qed.
Proof. intro n; induction n as [| p IHp]. - reflexivity. - simpl; rewrite <- IHp at 1; reflexivity. Qed.
Proof. intro n; induction n as [| p IHp]. - reflexivity. - simpl; rewrite <- IHp; auto with arith. Qed.
Proof. intro n; induction n as [| p IHp]. - reflexivity. - simpl; rewrite <- IHp; auto. Qed.
Proof. intro n; induction n as [| p IHp]. - reflexivity. - simpl; rewrite <- IHp; reflexivity. Qed.
Proof. intro n; induction n as [|n IH]. - reflexivity. - simpl; rewrite <- IH. reflexivity. Qed.
Proof. intro n; induction n as [|n IHn]. - reflexivity. - simpl; auto. Qed.
Proof. intro n; induction n as [|n IHn]. - reflexivity. - simpl; rewrite <- IHn; reflexivity. Qed.
Proof. intro n; induction n as [|n IHn]; simpl; auto. Qed.
Proof. intro n; induction n as [|n' IHn']. - reflexivity. - simpl; rewrite <- IHn'; reflexivity. Qed.
Proof. intro n; replace 0 with (S 0 - 1); auto. Qed.
Proof. intro n; rewrite (plus_comm n 0); auto with *. Qed.
Proof. intro n; rewrite (plus_comm n 0); auto with arith. Qed.
Proof. intro n; rewrite (plus_comm n 0); auto. Qed.
Proof. intro n; rewrite (plus_comm n 0); reflexivity. Qed.
Proof. intro n; rewrite (plus_comm n 0); simpl; auto. Qed.
Proof. intro n; rewrite (plus_comm n 0); trivial. Qed.
Proof. intro n; rewrite <- (plus_n_O n); reflexivity. Qed.
Proof. intro n; rewrite <- Nat.add_0_r at 1. reflexivity. Qed.
Proof. intro n; rewrite Nat.add_0_r; reflexivity. Qed.
Proof. intro n; rewrite Nat.add_comm; auto. Qed.
Proof. intro n; rewrite Nat.add_comm; reflexivity. Qed.
Proof. intro n; rewrite add_comm. reflexivity. Qed.
Proof. intro n; rewrite add_comm; auto with arith. Qed.
Proof. intro n; rewrite add_comm; reflexivity. Qed.
Proof. intro n; rewrite plus_O_r; reflexivity. Qed.
Proof. intro n; rewrite plus_comm with (m := 0); auto. Qed.
Proof. intro n; rewrite plus_comm; apply plus_O_n. Qed.
Proof. intro n; rewrite plus_comm; auto. Qed.
Proof. intro n; rewrite plus_comm; exact (plus_O_n n). Qed.
Proof. intro n; rewrite plus_comm; reflexivity. Qed.
Proof. intro n; simpl. auto with arith. Qed.
Proof. intro n; simpl; auto with arith. Qed.
Proof. intro n; simpl; auto. Qed.
Proof. intro; apply eq_add_S ; auto with arith. Qed.
Proof. intro; apply eq_add_S; auto. Qed.
Proof. intro; apply nat_ind with (P := fun n => n = n + O). - reflexivity. - intros; simpl; auto. Qed.
Proof. intro; apply sym_eq; apply Nat.add_0_r. Qed.
Proof. intro; elim n using nat_rect. - reflexivity. - intros m Hrec; simpl in |- *; rewrite Hrec; auto with arith. Qed.
Proof. intro; elim n. - reflexivity. - intros; simpl; auto with arith. Qed.
Proof. intro; elim n; simpl; auto. Qed.
Proof. intro; induction n as [| n IH]; simpl; auto. Qed.
Proof. intro; induction n as [| n IHn]. - reflexivity. - simpl; rewrite <- IHn; reflexivity. Qed.
Proof. intro; induction n as [| p IHp]; simpl; auto with arith. Qed.
Proof. intro; induction n as [| p IHp]; simpl; auto. Qed.
Proof. intro; induction n as [|n IH]; simpl; auto. Qed.
Proof. intro; induction n; simpl; auto. Qed.
Proof. intro; rewrite <- (Nat.add_comm 0); reflexivity. Qed.
Proof. intro; rewrite <- (add_comm 0); apply plus_O_n. Qed.
Proof. intro; rewrite <- (plus_n_O n); reflexivity. Qed.
Proof. intro; rewrite <- add_comm; auto with arith. Qed.
Proof. intro; rewrite <- add_comm; auto. Qed.
Proof. intro; rewrite <-plus_n_O. reflexivity. Qed.
Proof. intro; rewrite Nat.add_0_r; reflexivity. Qed.
Proof. intro; rewrite Nat.add_comm; auto with arith. Qed.
Proof. intro; rewrite Nat.add_comm; auto. Qed.
Proof. intro; rewrite Nat.add_comm; reflexivity. Qed.
Proof. intro; rewrite Nat.add_comm; trivial. Qed.
Proof. intro; rewrite add_comm; auto with *. Qed.
Proof. intro; rewrite add_comm; auto with arith. Qed.
Proof. intro; rewrite add_comm; auto. Qed.
Proof. intro; rewrite add_comm; reflexivity. Qed.
Proof. intro; rewrite add_comm; simpl; auto. Qed.
Proof. intro; rewrite add_comm; simpl; reflexivity. Qed.
Proof. intro; rewrite plus_comm; apply plus_O_n. Qed.
Proof. intro; rewrite plus_comm; auto with arith. Qed.
Proof. intro; rewrite plus_comm; auto. Qed.
Proof. intro; rewrite plus_comm; reflexivity. Qed.
Proof. intro; rewrite plus_comm; simpl; reflexivity. Qed.
Proof. intro; ring. Qed.
Proof. intro; ring_simplify; reflexivity. Qed.
Proof. intro; ring_simplify; trivial. Qed.
Proof. intro; simpl; auto with arith. Qed.
Proof. intros n; now rewrite <- Nat.add_comm. Qed.
Proof. intros. rewrite <- Nat.add_comm. apply eq_refl. Qed.
Proof. intros. rewrite <- plus_comm. reflexivity. Qed.
Proof. intros. rewrite Nat.add_comm; reflexivity. Qed.
Proof. intros; now rewrite Nat.add_0_r. Qed.
Proof. intros; now rewrite Nat.add_comm. Qed.
Proof. intros; rewrite <- (plus_n_O n); reflexivity. Qed.
Proof. intros; rewrite <- Nat.add_comm; reflexivity. Qed.
Proof. intros; rewrite <- add_comm; reflexivity. Qed.
Proof. intros; rewrite <- plus_n_O; reflexivity. Qed.
Proof. intros; rewrite Nat.add_comm; apply Nat.add_0_l. Qed.
Proof. intros; rewrite Nat.add_comm; apply add_O_l. Qed.
Proof. intros; rewrite Nat.add_comm; apply plus_O_n. Qed.
Proof. intros; rewrite Nat.add_comm; reflexivity. Qed.
Proof. intros; rewrite plus_comm; exact (plus_O_n n). Qed.
Proof. intros; ring. Qed.
Proof. intros; simpl; auto with arith. Qed.
Proof. simpl. auto with arith. Qed.
Proof. simple induction n; auto. Qed.
Proof. simple induction n; simpl in |- *; auto with arith. Qed.
Proof. simple induction n; simpl; auto with arith. Qed.
Proof. simple induction n; simpl; auto. Qed.
\end{lstlisting}

\phantomsection
\Urlmuskip=0mu plus 1mu\relax
\printbibliography

\end{document}